\documentclass{llncs}

\usepackage{hyperref}
\usepackage{color}
\usepackage{cleveref}
\usepackage{dsfont}
\usepackage{graphicx}
\usepackage{listings}
\usepackage{xcolor}
\usepackage[perpage]{footmisc}
  \newif\ifdraft
  \drafttrue 

\ifdraft
\fi

\newcommand{\name}[1]{\mathit{#1}}

\usepackage{xspace}
\newcommand{\jt}{\textsc{JaTeCS}\xspace}

\definecolor{light_gray}{rgb}{0.95,0.97,0.97}
\begin{document}

\title{\jt: an open-source JAva TExt Categorization System}

\author{Andrea Esuli, Tiziano Fagni, Alejandro Moreo Fern\'{a}ndez }

\institute{Istituto di Scienza e Tecnologie dell'Informazione \\
Consiglio Nazionale delle Ricerche, 56124 Pisa, IT\\
\email{first.second@isti.cnr.it}
}


\maketitle

\begin{abstract}
\jt is an open source Java library that supports research on automatic text categorization and other related problems, such as ordinal regression and quantification, which are of special interest in opinion mining applications.
It covers all the steps of an experimental activity, from reading the corpus to the evaluation of the experimental results. 
As \jt is focused on text as the main input data, it provides the user with many text-dedicated tools, e.g.: data readers for many formats, including the most commonly used text corpora and lexical resources, natural language processing tools, multi-language support, methods for feature selection and weighting, the implementation of many machine learning algorithms as well as wrappers for well-known external software (e.g., $\name{SVM_{light}}$) which enable their full control from code.
\jt support its expansion by abstracting through interfaces many of the typical tools and procedures used in text processing tasks.
The library also provides a number of ``template'' implementations of typical experimental setups (e.g., train-test, k-fold validation, grid-search optimization, randomized runs) which enable fast realization of experiments just by connecting the templates with data readers, learning algorithms and evaluation measures.
\end{abstract}


\section{Introduction}
The JAva TExt Categorization System (\jt) has been developed in the past years by our research group as the main software tool to perform research on a broad range of automatic text categorization \cite{sebastiani_machine_2002} and other text mining problems.

Research on text mining has gained new momentum in the last decade as the explosion of Social Networks (SNs\footnote{Here we broadly mean any platform that acts as a large-scale gatherer of user-generated content, thus ranging from Twitter to TripAdvisor, from Facebook to Amazon's user feedback.}) largely increased the amount of textual data generated on the Web and also accelerated the speed at which it is generated and consumed.
This scenario asked for novel methods to effectively and efficiently process such huge and novel streams of information, to sift down the relevant information with respect to a specific information need.
Moreover, the textual content generated in SNs is rich of information related to personal opinions and subjective evaluations, which are of great practical and commercial interest.
This aspect led to the growth of new disciplines such as Sentiment Analysis and Opinion Mining (SAOM, \cite{liu_sentiment_2012}). 
SAOM methods transform into a structured form the unstructured opinion-related information expressed by means of natural language in text, enabling the successive application of data mining methods on the structured information.
This transformation can be performed by processing text at various levels: processing each document as a single and atomic entity (e.g., classification), performing aggregated analysis on entire collections (e.g.,  quantification), extracting multiple piece of information for a document (e.g., information extraction).

\jt mainly consists of a Java library that implements many of the tools needed by a researcher to perform experiments on text analytics problems.
It covers all the steps and components that are usually part of a complete text mining pipeline: acquiring input data, processing text by means of NLP/statistical tools, converting it into a vectorial representation, applying optimizations in the vector space (e.g., projections into smaller space through matrix decomposition methods), application of machine learning algorithms (and the optimization of their parameters), and the application of the learned models to new data, evaluating their accuracy.
The development of \jt started with a specific focus on topic-oriented text categorization, but soon expanded toward many other text-categorization related problem, including sentiment analysis, transfer learning, distributional language modeling and quantification.

\jt is published at \url{https://github.com/jatecs/jatecs}, and it is released under the GPL v3.0 license\footnote{https://www.gnu.org/copyleft/gpl.html}.
In the following we describe the design concepts that drove the development of \jt (Section \ref{[jatecs:design_concepts]}), the core components (Section \ref{sec:core}, and how they can be combined to solve different text mining problems (Section \ref{sec:apps}). We conclude comparing \jt with similar software currently available (Section \ref{sec:related}).


\section{Design concepts}
\label{[jatecs:design_concepts]}

In an input dataset suitable for text analysis tasks, it is possible to describe its data by using 3 logical concepts and the relations among them.
The \textit{documents} are a logical representation of text instances that need to be processed.
A single document is composed by a set of \textit{features} where a single feature can be described by some attributes, e.g. a counter of the number occurrences of that feature in a specific document.
Moreover, in supervised learning problems, a document can have attached to it a set of \textit{labels} which best describe the content of that document in a custom taxonomy used in that specific problem.
Given these three foundational concepts it is possible to combine them together to model the relations between them that allow to answer questions like "which are all the documents that contain the feature \textit{x}?" or "Which are all the documents that contain feature \textit{x} and belong to label \textit{y}?".
The relation that pairs documents with features is the one that define the actual content of each document (if seen focusing on a document) and how a feature distributes among documents (if seen focusing on a feature). 
The relation that pairs documents with labels models the actual classification of documents in the taxonomy of interest.
The relation that pairs features with labels models how features distribute with respect to labels, and this information is the key of any supervised processing applied to data, from feature selection to the actual learning.

\begin{figure}[h!]
	\caption{Logical structure of a Jatecs index.}
	\centering
	\includegraphics[width=0.48\textwidth]{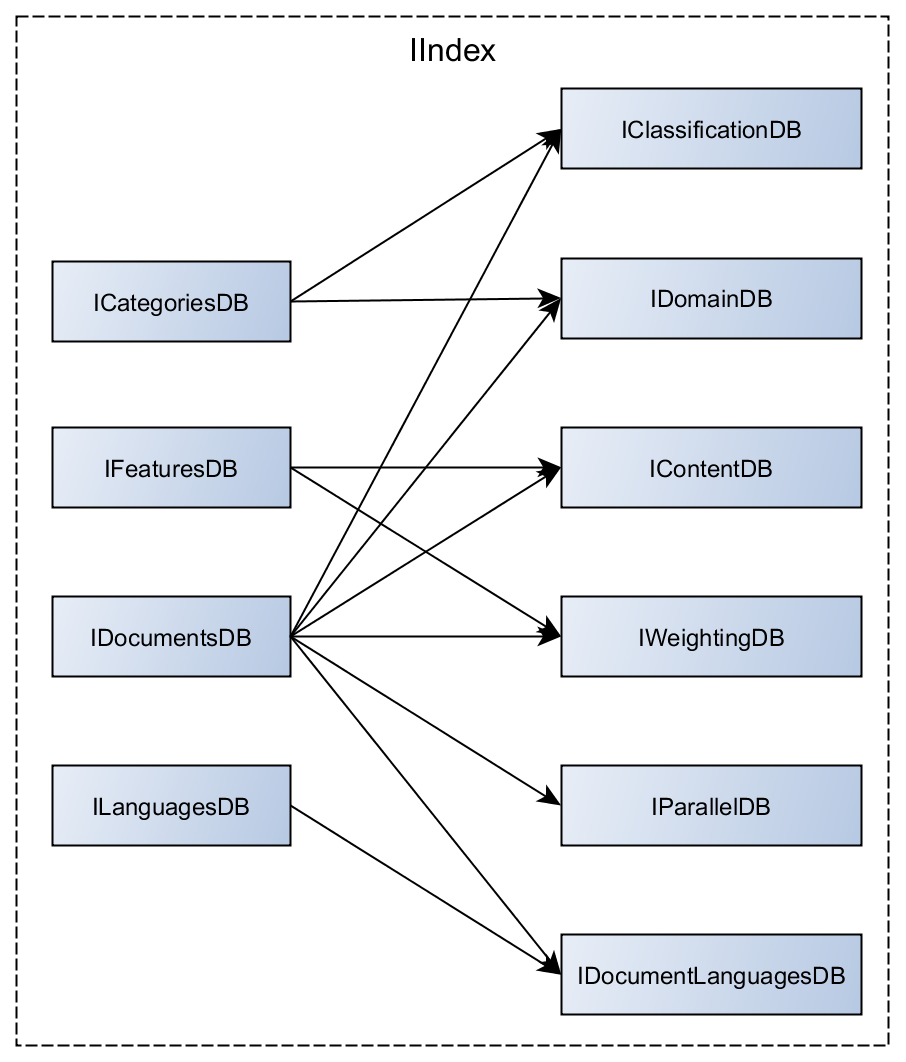}
	\label{fig:index}	
\end{figure}

All these concepts and relation are modeled in \jt by an API that defines each of them as an object with an interface to query information from it.
The design of code in \jt is corpus-centric: it is possible to trace back any feature, vector, label to the original corresponding entity in the corpus, in any moment of the computation. 
\jt core interface is named \textit{IIndex} and it keeps track of all data relations defined on a dataset (see Figure \ref{fig:index}).

The \textit{IIndex} is a data container which gives a unified access to views of the above mentioned concepts and relations, plus additional information that is relevant to some text mining problems (e.g., language of each document) or to some machine learning processes (e.g., feature weighting).
An IIndex is linked to three concept DBs (\textit{ICategoriesDB}, \textit{IFeaturesDB} and \textit{IDocumentsDB}) which contain information about the basic concepts of categories (labels), features and documents, and three relation DBs (\textit{IClassificationDB}, \textit{IContentDB} and \textit{IDomainDB}) which keep track of the existing relations among the basic concepts.
Additionally, \jt brings support to multilingual collections, through the concept DB \textit{ILanguagesDB}, and the relation DBs \textit{IParallelDB} and \textit{IDocumentLanguagesDB}.
In the following we are giving a more detailed description of each DB. 
\begin{description}
	
	\item[ICategoriesDB] This DB stores information about the set of all categories managed by the index. 
	Each category in the DB is represented by a pair \verb+<cID,categoryLabel>+ where \verb+cID+ is a category unique ID assigned by the system and \verb|categoryLabel| is the original string label associated with that category.
	\item[IFeaturesDB] This DB stores information about the set of all features managed by the parent index. 
	Each feature is represented by a pair \verb+<fID,featureText>+ where \verb+fID+ is a feature unique ID assigned by the system and \verb|featureText| is the original text associated with this feature and extracted from original dataset.
	\item[IDocumentsDB] This DB stores information about all the available documents managed by the parent index. 
	Each document in the DB is represented by a pair \verb|<dID,docName>| where \verb|dID| is the document unique ID assigned by the system and \verb|docName| is the document logical name linked to external input dataset.
	\item[ILanguagesDB] This DB stores information about all the available languages managed by the parent index. Each language in the DB is represented by a unique predefined label, e.g., \emph{en}, \emph{it}, or \emph{es}, for English, Italian, or Spanish languages, respectively.
	\item[IClassificationDB] This sparse DB stores the relations existing in the input dataset between  the categories and the documents, i.e. it links together the  ICategoriesDB and the IDocumentsDB. The DB is able to handle multilabel contexes, so for each document it is possible to assign any numbers of labels in the form of list of pairs \verb|<dID,cID>|. Here \verb|dID| and \verb|cID| have the usual meaning as described above.
	\item[IContentDB] This sparse DB stores the relations existing in the input dataset between the documents and the features, i.e. it links together the IDocumentsDB and the IFeaturesDB. The presence of a specific feature (having \verb|fID| identifier) in a given document (having \verb|dID| identifier) is marked in this DB by keeping track the number of occurrences of the feature in the document. If a feature does not appear in a document then the DB will not store any entry for the considered feature and document.
	\item[IDomainDB]  This sparse DB stores the relations existing in the input dataset between the features and the categories, i.e. it links together the IFeaturesDB and the ICategoriesDB. Normally the index has a global representation of the features in all available categories, so each feature is valid in all categories. Instead sometime it is useful to give to each feature a specific valid categories context, i.e. decide if a feature (having \verb|fID| identifier) is or not valid in a specific category (having \verb|cID| identifier) \footnote{This configuration is also called local representation of the features.}. In this last case, this DB facilitates this task by allowing to keep track of the validity of features inside each specific category.
	\item[IWeightingDB] This DB is similar to IContentDB because stores the relations between the documents and the features. Differently from IContentDB, it allows to keep track of a weight associated to a feature in a specific document. Usually machine learning algorithms use the feature weights to build the learning model. As we will see in section \ref{jatecs:dimensionalityReduction}, \jt offers several methods to perform the weighting process over the features contained in an index.   
	\item[IDocumentLanguagesDB] This DB links together the IDocumentsDB with the ILanguagesDB, that is, it allows for specifying the language or languages in which a given document is written. This DB might only exist in multilingual indices, and could be obviated for monolingual ones.
	\item[IParallelDB] This DB allows to store relations of parallelism among documents. Each relation is represented as a tuple \verb|<dID_0,dID_1,...>|, indicating documents with ID \verb|dID_0|, \verb|dID_1|, ... consist of parallel versions about the same content\footnote{Whether the sense of \emph{parallelism} stands for sentence-level, document-level, or topic-level is an implementation decision.}.
\end{description}
The IIndex implementation currently available in \jt is suitable for completely in-memory data processing. During initial data loading of an input dataset, the programmer can specify how to store data for \emph{IClassificationDB} and \emph{IContentDB}, choosing for these DBs a trade-off between RAM memory consumption and efficiency in the access of their data. Anyway, thanks to the extensive use of interfaces and abstract classes, a different IIndex implementation (e.g. disk-based with an effective caching mechanism) could be realized with little effort.\\
Following is a code example of usage of a \jt index:\\\\
\begin{minipage}{\linewidth}
\begin{lstlisting}[language=Java, basicstyle=\footnotesize\ttfamily]
IIndex index = .... // Built from some input data source.

// Iterate over all documents.
IIntIterator documents = index.getDocumentDB().getDocuments();
while(documents.hasNext()) {
	int docID = documents.next();
	String documentName = index.getDocumentDB().getDocumentName(docID);
	(continues on next page)
	\end{lstlisting}
\end{minipage}\\\\
\begin{minipage}{\linewidth}
	\begin{lstlisting}[language=Java, basicstyle=\footnotesize\ttfamily]	
	(continues from previous page)
	// Select only the features which have number of occurrences >= 5 or weight > 0.15.
	ArrayList<String> wantedFeatures = new ArrayList<>();
	IIntIterator features = index.getContentDB().getDocumentFeatures(docID);
	while(features.hasNext()) {
		int featID = features.next();
		int frequency = index.getContentDB().getDocumentFeatureFrequency(docID, featID);
		double weight = index.getWeightingDB().getDocumentFeatureWeight(docID, featID);

		if (frequency >= 5 || weight > 0.15)
			wantedFeatures.add(index.getFeatureDB().getFeatureName(featID));
	}

	// Print information about the document.
	System.out.println("****************************");
	System.out.println("Document name: "+documentName);
	System.out.println("Most important features: "+Arrays.toString(wantedFeatures.toArray()));
}	
\end{lstlisting}
\end{minipage}


\section{Core library}\label{sec:core}

In this section we report on the various components that are implemented in \jt to support the development of a complete text processing pipeline.

\subsection{Data formats and corpora support}
\jt allows to import text from various data sources and different formats.
All the classes related data input (and output, i.e., saving back a IIndex to a specific format) are defined in the \texttt{jatecs.indexing.corpus} namespace.
The abstract class \texttt{CorpusReader} defines the core methods to access a corpus, i.e., accessing its files, selecting the relevant set of documents (e.g., when there are well-known train/test splits), returning an iterator on its documents and their labels.
\jt provides specializations of the \texttt{CorpusReader} to quickly import data in different formats and have it available as a standard IIndex structure. 
This index can then be used in all subsequent stages of the experimentation and manipulated by the algorithms provided by the library, e.g., to perform a feature selection operation or to build a classifier.

The library implements readers for the following corpora:
\begin{description}
	\item[Reuters21578]\footnote{http://www.daviddlewis.com/resources/testcollections/reuters21578/} Currently it is probably the most widely used test collection for text categorization research. 
	This dataset defines several data splits to be used for experimentation, the most used is the ``ModApte'' split which has 9,603 documents in training set, 3,299 documents in test set and a taxonomy consisting in 115 categories.
	\item[RCV1-v2/RCV2] RCV1-v2\cite{RCV1} is another popular text
	categorization benchmark made available by Reuters and consisting of 804,414
	news stories produced by Reuters from 20 Aug 1996 to 19 Aug 1997. 
	The library provides access to ``LYRL2004'' data split, consisting in 23,149 training documents, 781,265 test documents and a taxonomy of 103 categories.
	RCV2\footnote{http://trec.nist.gov/data/reuters/reuters.html} is a multilingual extension corpora containing over 487,000 news stories collected in the same time frame in thirteen different languages other than English (Dutch, French, German, Chinese, Japanese, Russian, Portuguese, Spanish, Latin American Spanish, Italian, Danish, Norwegian, and Swedish). 
	RCV2 is a corpus comparable at topic level.
	\item[WipoAlpha]\footnote{http://www.wipo.int/classifications/ipc/en/ITsupport/Categorization/dataset/wipo-alpha-readme.html} It is a large collection published by the World Intellectual Property Organization
	(WIPO) in 2003. 
	The dataset consists of 75,250 patents classified according to version 8 of the International Patent Classification scheme and a hierarchical taxonomy of more than 69,000 codes.
	\item[Oshumed]\footnote{http://davis.wpi.edu/xmdv/datasets/ohsumed.html} The OHSUMED test collection is a set of 348,566 references from MEDLINE, the on-line medical information database, consisting of titles and/or abstracts from 270 medical journals over a five-year period (1987-1991). 
	The data is classified into a taxonomy of 23 cardiovascular diseases categories.
	\item[JRCAcquis] JrcAcquis\cite{jrcAcquis} is a parallel corpora available in more than 20 languages containing approximately 20,000 documents per language about legislative texts from EU legislation. 	
	The data is classified manually using the Eurovoc thesaurus, which consists of over 6,000 descriptor terms (classes) organized hierarchically into up to eight levels.
	\item[20 Newsgroup]\footnote{http://qwone.com/~jason/20Newsgroups/} another very popular collection of approximately 20,000 documents partitioned nearly evenly in 20 different Usenet discussion groups. 
	\item[Multi-Domain Sentiment Dataset v.2]\footnote{http://www.cs.jhu.edu/~mdredze/datasets/sentiment/} \cite{Blitzer:2007gf}, contains product reviews from Amazon.com for four domains (Books, DVDs, Electronics, and Kitchen appliances) and is commonly used as a benchmark for domain adaptation techniques. 
	The dataset comprises 1000 positive reviews 
	and 1000 negative reviews 
	for each domain, and a set of unlabelled documents ranging from 3,586 to 5,945 per domain.
	
	\item[Webis-CLS-10]	\footnote{http://www.uni-weimar.de/en/media/chairs/webis/research/corpora/corpus-webis-cls-10/} \cite{Prettenhofer:2010ys} is a cross-lingual sentiment collection consisting of Amazon product reviews written in four languages
	(English, German, French, and Japanese), covering three product domains (Books, DVDs, Music).
	For each language-domain pair there are 2,000 training documents, 2,000 test documents, and from 9,000 to 50,000 unlabelled documents.
\end{description}

The library also implements readers for the following common formats:

\begin{description}
	\item[SvmLight/LibSVM] SvmLight\cite{Joachims:1999} is a popular software providing Support Vector Machines (SVMs) algorithms for classification and regression problems. 
	The software defines a standard format to manipulate input data and this is used also by LibSVM\cite{libSVM}, another popular package used to perform SVM classification. 
	A lot of datasets are already available in this specific data format\footnote{https://www.csie.ntu.edu.tw/~cjlin/libsvmtools/datasets/}. \jt support this format and make it easy to work with such type of data.
	\item[CSV] This is the classic Comma Separated Values file format. 
	The user can specify the character value to use as a separator and the data can be available in multiple files.
	\item[ARFF] (Attribute-Relation File Format)\footnote{http://weka.wikispaces.com/ARFF} is a file format to describe instances of entities (documents, in our case) that share a set of attributes (features, in our case).
\end{description}


\subsection{Text processing and feature extraction}
All the methods that typically compose the processing pipeline that converts text for strings of characters to vectors in a vectorial space are modeled through interfaces or abstract classes.
Each interface/abstract class has then a number of implementations for the most commonly used methods.
For example, the basic activities of the process of feature extraction are implemented in the abstract class \texttt{FeatureExtractor}: stop word removal, stemming (supporting various languages through the use of Porter stemmers generated by Snowball\cite{Snowball} software), substitution of xml/html entities with proper characters.
The \texttt{FeatureExtractor} is then extended to cover various cases:
\begin{description}
	\item[BOWFeatureExtractor] This is a basic extractor that splits of a given text content using not only spaces but also all the common punctuation signs and parenthesis like "\verb|,|","\verb|.|", "\verb|;|", "\verb|(|", etc.
	\item[CharsNGramsFeatureExtractor] This extractor generates n-grams of characters (with $n$ specifiable by the user) from a given text content. N-grams can be word-bounded or extracted continuously from the text string, regardless of word tokenization.
	\item[NLPFeatureExtractor] This extractor use the Stanford POS tagger \cite{manning2014stanford} and sentiment lexica \cite{baccianella2010sentiwordnet,bloom2007extracting,stone1966general} to extract sentiment-relevant features. The extractor marks the presence of terms with known sentiment polarity from sentiment lexica and also generates multi-word feature using POS patterns and negation propagation methods (as described in \cite{baccianella2009multi}).
	\item[SetFeatureExtractor] This is an aggregator that allows to use a set of other  extractors to combine the features generated by them, optionally marking such feature in a way that eventual identical features from different extractors are considered as different features (e.g., the word ``and'' from BOWFeatureExtractor is not confused with the 3-gram ``and'' from CharsNGramsFeatureExtractor).
\end{description}
Following is a \jt code excerpt of loading an input data source using a CSV reader and performing feature extraction using the CharsNGramsFeatureExtractor:\\\\
\begin{minipage}{\linewidth}
	\begin{lstlisting}[language=Java, basicstyle=\footnotesize\ttfamily]
	// Load all labels (categories) of interest.
	TroveCategoryDBBuilder categoryDBBuilder = new TroveCategoryDBBuilder();
	FileCategoryReader categoriesReader = new FileCategoryReader(categoriesFile,
		categoryDBBuilder);
	ICategoryDB categoryDB = categoriesReader.getCategoryDB();
	
	// Prepare a character n-grams feature extractor.
	CharsNGramFeatureExtractor extractor = new CharsNGramFeatureExtractor();
	extractor.enableStemming(new EnglishPorterStemming());
	extractor.enableStopwordRemoval(new EnglishStopword());
	extractor.setNGramSize(4);
	
	// Prepare a CSV corpus reader.
	CSVCorpusReader corpusReader = new CSVCorpusReader(categoryDB);
	corpusReader.setFieldSeparator("\t");
	corpusReader.setInputFile(dataFile);
	corpusReader.setDocumentSetType(SetType.TRAINING);
	
	// Build the index.
	TroveMainIndexBuilder mainIndexBuilder = new TroveMainIndexBuilder(
	categoryDB);
	FullIndexConstructor mainIndexConstructor = new FullIndexConstructor(
		corpusReader, mainIndexBuilder);
	mainIndexConstructor.setFeatureExtractor(extractor);
	mainIndexConstructor.exec();
	IIndex index = mainIndexConstructor.index();
	\end{lstlisting}
\end{minipage}


\subsection{Dimensionality Reduction}
\label{jatecs:dimensionalityReduction}

The number of different features resulting from the indexing process depends almost directly on the number of different terms in a corpus.
It is typically a very high number with respect to the size of feature sets from data sources other than text, which can be a performance issue for many learning algorithms.
Another typical difference from another media data sources, is that features follow a power-law distribution, i.e., few features appears a lot of time and a lot of features appear few times in text.
For example, the relatively small set of 9,000 documents of the Reuters21578 dataset are composed by 20,123 distinct words (number excluded), but only 313 of them determine half of the half million word occurrences of that dataset.
Features at the extremes of the distribution are typical less informative than those in the center, but it is not obvious which feature can be discarded after the feature extraction process. 

Dimensionality Reduction \cite{sebastiani_machine_2002} techniques aim at reducing the number of dimensions of the feature space by selecting the most informative features.
There are mainly two approaches for reducing the dimensionality: Feature Selection and Distributional Semantic Models, and \jt supports both of them.

\textbf{Feature Selection (FS)} methods attempt to select a reduced subset of informative features from the original set (thus discarding the rest), so that the size of the new subset is much smaller than the original one and so that the reduced set yields high classification effectiveness. 
In Text Classification (TC) the problem is usually tackled via ``filtering'' approaches, i.e., methods relying on a mathematical function measuring the contribution of each feature to the classification task. 
\jt implements a number of \emph{filtering} approaches for FS (see e.g., \cite{yang1997comparative}), covering most popular Term Space Reduction (TSR) functions, including \emph{Information Gain}, $\chi^2$, \emph{Pointwise Mutual Information}, or \emph{Odds Ration}, among many others; as well as a number of well-known TSR policies, including the \emph{Local} selection for each category, the \emph{max}, \emph{sum}, and \emph{weighted} variants of the \emph{Global} selection for all categories, and the \emph{Round Robin} \cite{FormanRoundRobin} policy.
The following snippet shows a use information gain combined with round robin in \jt:\\\\
\begin{minipage}{\linewidth}
	\begin{lstlisting}[language=Java, basicstyle=\footnotesize\ttfamily]
		//applies Round Robin policy to select the 5000 most important features according to Information Gain
		RoundRobinTSR rrTSR = new RoundRobinTSR(new InformationGain());
		rrTSR.setNumberOfBestFeatures(5000);
		rrTSR.computeTSR(index);
	\end{lstlisting}
\end{minipage}

\textbf{Distributional Semantic Models (DSM)} transform the original feature space into a new space of reduced dimensionality where semantic between terms is explicitly modeled thorough the concept of \textit{distance} (between pairs of terms) and \textit{position} in the space. 
Dimensions in a DSM are no longer associated to one particular term, as in a traditional bag of words model, but to \textit{latent} features.
\jt implements a number of DSM, supporting Latent Semantic Analysis\footnote{As a wrapper of SVDLIBC, see \url{http://tedlab.mit.edu/~dr/SVDLIBC/}} \cite{deerwester1990indexing} and most relevant Random Projections approaches \cite{kaski1998dimensionality} such as Random Indexing \cite{kanerva2000random,sahlgren2005introduction}, Lightweight Random Indexing \cite{LriJair2016}, or the Achlioptas mapping \cite{achlioptas2001database}.
The following snippet shows the use of random indexing in \jt:\\\\
\begin{minipage}{\linewidth}
	\begin{lstlisting}[language=Java, basicstyle=\footnotesize\ttfamily]
		// Use Random Indexing to map the original feature space into a 5000-dimensional one distributing 1% of non-zero values for each random index
		int dim=5000;
		RandomIndexing randomIndexing = new RandomIndexing(index, dim, dim*0.01);
		randomIndexing.project();
		index = randomIndexing.getLatentTrainindex();
	\end{lstlisting}
\end{minipage}

\subsection{Feature Weighting}
\label{jatecs:featureWeigthing}
Feature frequency (as reflected in \textit{IContentDB}) may not suffice as an indicator of the importance of a term to the document content (e.g., Na\"ive Bayesian learner uses frequencies, SVM work better with a proper weighting).

For this reason \jt maintains a dedicated data structure, the \textit{IWeightingDB}, to explicitly quantify the relative importance of a term to a document thorough real values.
The Feature Weighting functions are responsible of setting those values, and \jt implements most widely used weighting criteria, including the well-known \emph{tf-idf}\cite{Salton:1988} (and many of its variants) and \emph{BM25}\cite{bm25}.
The following snippet exemplifies how to apply the normalized \emph{tf-idf} to bring to bear the terms importance to the documents.\\\\
\begin{minipage}{\linewidth}
	\begin{lstlisting}[language=Java, basicstyle=\footnotesize\ttfamily]
	// Weight the features using TF-IDF.
	IWeighting weightingFunction = new TfNormalizedIdf(index);
	index = weightingFunction.computeWeights(index);
	// now index has a IWeightingDB with tfidf weights
	\end{lstlisting}
\end{minipage}

\subsection{Learning algorithms}
\jt implements many machine learning algorithms for classification, ordinal regression, and clustering.  
All those algorithms implement shared interfaces that enable the programmer to easily test different learning methods.
Both interfaces and algorithms are defined in the \texttt{it.cnr.jatecs.classification} namespace.
A learning algorithm must extend the \texttt{BaseLearner} abstract class by implementing the method\\\\
\begin{minipage}{\linewidth}
	\begin{lstlisting}[language=Java, basicstyle=\footnotesize\ttfamily]
	public abstract IClassifier build(IIndex trainingIndex);
	\end{lstlisting}
\end{minipage}
that, given an IIndex return an IClassifier, which in turns requires the implementation of the methods\\\\
\begin{minipage}{\linewidth}
	\begin{lstlisting}[language=Java, basicstyle=\footnotesize\ttfamily]
	public ClassificationResult classify(IIndex testIndex, int document);
	public ClassificationResult[] classify(IIndex testIndex, short category);
	\end{lstlisting}
\end{minipage}

i.e., a method to classify a single document in a IIndex with respect to all the possible labels, and a method to classify all the documents in a IIndex with respect to a single label\footnote{These two methods can be use interchangeably to obtain a classification of all documents in an IIndex with respect to all possible labels. The choice of which one to use is an optimization parameter that can be set depending on how the used classifier works best. For example, SVMs works better classifying category by category, while AdaBoost works better classifying document by document.}.

Regarding classification, \jt covers the most popular learning methods such as the Na\"{i}ve Bayes \cite{Lewis-naivebayes},  Rocchio, Logistic regression, Ridge regression \cite{ridgeregression}, k-NN \cite{Yang-knn}, AdaBoost and boosting methods in general \cite{Schapire:boosting}, and Support Vector Machines (SVMs). 
SVMs support includes wrappers for the popular SVM$^{\name{light}}$\footnote{\url{http://svmlight.joachims.org/}} \cite{Joachims:1999}, SVM$^{\name{perf}}$ \cite{Joachims05}, and LibSVM\footnote{\url{https://www.csie.ntu.edu.tw/~cjlin/libsvm/}} packages.
Classifiers committees (ensembles) and bagging methods are also implemented.

For ordinal regression, the library provides wrappers for $\epsilon$-SVR and $\nu$-SVR \cite{chu2007support}, as well as implementations of graph based methods, such as regression trees and D-DAGs \cite{platt1999large}.
A highly customizable implementation of $k$-means is implemented for clustering.

The following snippet offers an example on a basic text classification pipeline in \jt.\\\\
\begin{minipage}{\linewidth}
	\begin{lstlisting}[language=Java, basicstyle=\footnotesize\ttfamily]
	IIndex train = .... // Training data.
	IIndex test = ....  // Test data.
	
	ILearner learner = null;
	if(useBayes)
		learner = new NaiveBayesLearner();
	else
		learner = new SVMLearner();
	// the rest of the code is independent of the selected learner
	IClassifier	classifier = learner.build(train);
	
	// Classify test documents and get their predictions.
	Classifier classifierModule = new Classifier(test, classifier);
	classifierModule.exec(); 
	IClassificationDB predictions = classifierModule.getClassificationDB();
	\end{lstlisting}
\end{minipage}
\vspace{-0.1in}

\subsection{Evaluation}
\jt provides a standard set of tools to experimentally measure the effectiveness of a system built using one of the available supervised algorithms. In case of evaluation of a classifier, we need to keep track of all correct/wrong predictions made by it over a test documents set. As explained in \cite{sebastiani_machine_2002}, when evaluating a multi-label classification system, one can use multiple \emph{contingency tables} (one for each label is interested on) to compute various error measures like \emph{precision}, \emph{recall}, \emph{accuracy}, \emph{f1}, etc. \jt provides all these measures both at the level of a single label and in a aggregate way in order to \emph{micro/macro} evaluate \cite{sebastiani_machine_2002} the effectiveness of the built system as a whole. In the same way, the software is also able to handle hierarchical classification systems by evaluating in a special way all internal classifier nodes acting as non-terminal labels.\\
Multi-class single-label or binary classifiers can be instead evaluated through the use of a \emph{confusion matrix}, a structure relating all labels decisions and giving the exact repartition of wrong decisions for a specific label among all other labels. Our software offers a specific class to perform this task.\\
\jt is also able to evaluate a regression model by providing an evaluator which computes, for each label, a simple difference, in absolute value, between the expected value and value estimated by the regressor.\\
Here is a small snippet of code showing how to perform classifier evaluation:\\\\
\begin{minipage}{\linewidth}
	\begin{lstlisting}[language=Java, basicstyle=\footnotesize\ttfamily]
	IClassificationDB predictions = ... // Predictions made by a classifier.
	IClassificationDB goldStandard = ... // Real documents lables.
	
	// Evaluate classifier predictions.
	ClassificationComparer comparer = new ClassificationComparer(predictions, goldStandard);
	ContingencyTableSet tableSet = comparer.evaluate();
	
	// Print evaluation measures.
	ContingencyTable table = tableSet.getGlobalContingencyTable();
	System.out.println("Global results (micro-averaged evaluation)");
	System.out.println("tp = " + table.tp()
	+ "\ttn = " + table.tn()
	+ "\tfp = " + table.fp()
	+ "\tfn = " + table.fn());
	
	String res = String.format("p = %.3f\tr = %.3f\tf1 = %.3f\ta = %.3f",
		table.precision(), table.recall(),
		table.f1(), table.accuracy());
	System.out.println(res);
	\end{lstlisting}
\end{minipage}
\vspace{-0.1in}


\section{Applications}\label{sec:apps}

\subsection{Classification}

\jt handles the most common types of classification tasks \cite{sebastiani_machine_2002}, starting from simple binary classification, and including multi-label classification, multi-class single-label classification, and hierarchical classification. 
Multi-class, single-label classification is supported by the implementation of the \textit{one-vs-all} \cite{Bishop:ML}, \textit{one-vs-one} \cite{Bishop:ML}, and D-DAGs \cite{platt1999large,Rifkin_one_vs_all} methods, on which any binary algorithm can be plugged in.\\
The \jt index structure is able to manage taxonomies with a hierarchical structure, and implements a hierarchical classification learner \cite{Chakrabarti1998,Yang2003}, which can use any learning algorithm as the basic learning device. 
The hierarchical learner supports the customization of the selection policy for negative examples for each node of the hierarchy, implementing the \textit{Siblings}, \textit{All}, \textit{BestGlobal}, and \textit{BestLocal(k)} policies of \cite{Fagni_Negatives}.\\
Classification on multilingual (comparable or parallel) corpora \cite{Adeva:2005zx} is supported, including an implementation of the Multilingual Domain Models \cite{gliozzo2005cross} technique.
Optimization of parameters is supported by the implementation of well known exploration/validation methods, including \textit{K-Fold cross validation} (simple or stratified) \cite{Kohavi1995}, \textit{leave-one-out} \cite{Kohavi1995} and \textit{grid-search} \cite{Bengio2012}.\\ 
Following is an example of K-Fold cross evaluation using a TreeBoost classifier \cite{Esuli-TreeBoost}:\\\\
\begin{minipage}{\linewidth}
	\begin{lstlisting}[language=Java, basicstyle=\footnotesize\ttfamily]
	// Build base internal learner using MP-Boost.
	AdaBoostLearner internalLearner = new AdaBoostLearner();
	AdaBoostLearnerCustomizer internalCustomizer = new AdaBoostLearnerCustomizer();
	internalCustomizer.setNumIterations(iterations);
	internalCustomizer.setWeakLearner(
	new MPWeakLearnerMultiThread(threadCount));
	internalCustomizer
	.setInitialDistributionType(InitialDistributionMatrixType.UNIFORM);
	internalLearner.setRuntimeCustomizer(internalCustomizer);
	
	// Build treeboost using the specified internal learner.
	TreeBoostLearner learner = new TreeBoostLearner(internalLearner);
	TreeBoostLearnerCustomizer customizer = new TreeBoostLearnerCustomizer(
	internalCustomizer);
	learner.setRuntimeCustomizer(customizer);
	(continues on next page)
	\end{lstlisting}
\end{minipage}\\\\
\begin{minipage}{\linewidth}
	\begin{lstlisting}[language=Java, basicstyle=\footnotesize\ttfamily]	
	(continues from previous page)
	// Prepare k-fold evaluator.
	AdaBoostClassifierCustomizer internalClassifierCustomizer = new AdaBoostClassifierCustomizer();
	internalClassifierCustomizer.groupHypothesis(true);
	TreeBoostClassifierCustomizer classifierCustomizer = new TreeBoostClassifierCustomizer(
	internalClassifierCustomizer);
	SimpleKFoldEvaluator kFoldEvaluator = new SimpleKFoldEvaluator(learner,
	customizer, classifierCustomizer, true);
	kFoldEvaluator.setKFoldValue(10);
	
	// Evaluate the TreeBoost learner over the specified training data.
	IIndex training = ...
	ContingencyTableSet tableSet = kFoldEvaluator.evaluate(training, null);
	\end{lstlisting}
\end{minipage}
\vspace{-0.1in}

\subsection{Active learning, training data cleaning, and semi automated text classification}

\jt provides implementations of a rich number of methods proposed for three classes problems that are interrelated: Active Learning (AL), Training Data Cleaning (TDC), and Semi-Automated Text Classification (SATC).

In AL \cite{Hoi:2006ef,li2013active,Tong01} the learning algorithm can select which documents to add to the training set at each step of an iterative process. The aim is to minimize the amount of human labeling needed to obtain high accuracy.
\jt{} implements the svm-based AL method proposed in \cite{li2013active}: Max-Margin Uncertainty sampling (MMU), Label Cardinality Inconsistency (LCI).

The TDC task \cite{Esuli:2013ko,nakagawa2002detecting} consists of using learning algorithms to discover labeling errors in an already existing training set.
In AL such information is not available, and thus TDC can leverage on more information, and different strategies from AL, to identify training labeling errors that once corrected will contribute to improve the performance of the classifier.
\jt{} implements a number of training data cleaning policies described in \cite{Esuli:2013ko}, e.g., cleaning by confidence, by committee disagreement, by $k$-nn labeling similarity.

SATC \cite{Berardi:2012fk,Martinez2013} aims at reducing the amount of effort a human should invest while inspecting, and eventually repairing, the outcomes produced by a classifier in order to guarantee a required accuracy level.
SATC task differs from both AL and TDC in the fact that the goal is not improving a classifier but the accuracy of a classification of a collection for its use in an external task (e.g., an e-discovery process), where such external task may determine a different importance in correcting difference type of errors.
\jt{} implements SATC-oriented methods based on confidence ranking and utility ranking \cite{Berardi:2012fk}.

\subsection{Transfer learning}
Transfer learning \cite{pan2010survey} aims at enabling machine learning methods learn effective classifiers for a ``target" domain when the only available training data belongs to a different ``source" domain.
This problem could be posed as how to bring closer both data distributions.
Distances between domains are typically quantified as the \emph{$\mathcal{A}$-distance} \cite{Ben-David:2006hs}, a useful tool for transfer learning research that is incorporated in \jt. 
The framework also includes an implementation of the Distributional Correspondence Indexing (DCI) \cite{dciecir2015} algorithm for cross-domain and cross-lingual learning.
The following code example shows how a cross-domain classifier can be learned from a source domain and applied to a different target domain by using sets of unlabeled documents from both domain to define a common latent projection space, built using DCI methods.
Note how the portion of code that implements the DCI projection can be removed without affecting the functionality of the rest of the code.\\\\
\begin{minipage}{\linewidth}
	\begin{lstlisting}[language=Java, basicstyle=\footnotesize\ttfamily]
	//reading indexes
	IIndex train_s = ... // train documents from source domain
	IIndex test_t = ... // test documents from target domain
	
	// Start of DCI-related code
	IIndex unlabel_s = ... // unlabeled documents from source domain
	IIndex unlabel_t = ... // unlabeled documents from target domain
	
	// cosine-based DCF
	IDistributionalCorrespondenceFunction distModel_source= new CosineDCF(unlabel_s);
	IDistributionalCorrespondenceFunction distModel_target= new CosineDCF(unlabel_t);

	/* other DCF models available:  LinearDCF, PointwiseMutualInformationDCF, MutualInformationDCF, PolynomialDCF, GaussianDCF */
	
	// projecting indexes into DCF space
	DistributionalCorrespondeceIndexing dci = new DistributionalCorrespondeceIndexing(train_s, 
			distModel_source, distModel_target, customizer);
	dci.compute();
	train_s = dci.getLatentTrainIndex();
	test_t = dci.projectTargetIndex(test_t);
	// End of DCI-related code

	//train learner
	IClassifier classifier = Utils.trainSVMlight(train_s, svmconfig);
		
	(continues on next page)
	\end{lstlisting}
\end{minipage}\\\\
\begin{minipage}{\linewidth}
	\begin{lstlisting}[language=Java, basicstyle=\footnotesize\ttfamily]	
	(continues from previous page)

	//classification
	IClassificationDB predictions = classification(dci, classifier, test_t);

	//evaluation
	IClassificationDB trueValues = test_t.getClassificationDB();
	Utils.evaluation(predictions, trueValues, resultsPath, targetTestName);
\end{lstlisting}
\end{minipage}

\subsection{Quantification}
Quantification \cite{Forman:2008} is the problem of estimating the distribution of labels in a collection of unlabeled documents, when the distribution in the training set may substantially differ.
Though quantification processes a dataset as a single entity, the classification of single documents is the building block on which many quantification methods are built. 
\jt\ implements the three best performing classification-based quantification methods of \cite{Forman:2008} as well as their probabilistic versions, as proposed in \cite{bella2010quantification}. 
The implementation is independent from the underlying classification method, which acts as a plug-in component, as shown in the following sample code:

\begin{lstlisting}[language=Java, basicstyle=\footnotesize\ttfamily]
int folds = 50;
IScalingFunction scaling = new LogisticFunction();
// any other learner can be plugged in
ILearner classificationLearner = new SvmLightLearner();

// learns six different quantifiers on training data
QuantificationLearner quantificationLearner = new QuantificationLearner(
folds, classificationLearner, scaling);
QuantifierPool pool = quantificationLearner.learn(train);

// quantifies on test returning the six predictions
Quantification[] quantifications = pool.quantify(test);
// evaluates predictions against true quantifications
QuantificationEvaluation.Report(quantifications,test);
\end{lstlisting}
\vspace{-0.1in}

\subsection{Imbalanced Text Classification}
The accuracy of many classification algorithms is known to suffer when the data are imbalanced (i.e., when the distribution of the examples across the classes is severely skewed). Many applications of binary text classification are of this type, with the positive examples of the class of interest far outnumbered by the negative examples. Oversampling (i.e., generating synthetic training examples of the minority class) is an often used strategy to counter this problem. \jt provides a number of SMOTE-based implementations, including the original SMOTE approach \cite{chawla2002smote}, Borderline-SMOTE \cite{han2005borderline}, and SMOTE-ENN \cite{batista2004study}, and the probabilistic topic model called DECOM \cite{chen2011exploiting}. \jt also provides an implementation of the recently proposed Distributional Random Oversampling (DRO -- \cite{dro_sigir}) , an oversampling method specifically designed for classifying data (such as text) for which the distributional hypothesis holds.


\section{Related work}\label{sec:related}
On the market there are several software similar in spirit to \jt and optimized for data mining and NLP tasks. 
Some of these are open source, others are commercial products and/or closed source. 
In this section we will concentrate only on open source software because, being freely available to obtain and use, it is easier to test and evaluate the quality and the features of the software.

Waikato Environment for Knowledge Analysis (Weka)\cite{hall2009weka} is a very popular Java software suite of ML algorithms useful for generic data mining tasks.
It is probably one of the oldest open source data mining tool available in the market (first public Java version released in 1999) and, due to its free availability and the vast amount of available features, has gained a lot of traction in the past years, having being used extensively both in academic and industrial contexts. 
Weka, differently from \jt that is focused on textual data, supports several standard data mining tasks on several data types (e.g. numeric, nominal, etc.), by including algorithms for data preprocessing, feature selection, classification, regression, clustering and data visualizations.
The software uses the Attribute-Relation File Format (ARFF) as the default format used to ingest data from input datasets, although as \jt it supports other popular input data formats like CSV or LibSVM/SVMLight sparse representation.
All the algorithms provided by Weka assume that each input data source (dataset) is represented in terms of instances (documents), where each instance is composed by the same number of attributes and each attribute has its own type (e.g. an attribute named "age" has type "numeric").
A dataset can be manipulated through the use of filters, which allow to preprocess the data in some way (e.g. remove frequent/infrequent attributes, discretize numeric attributes, etc.) before passing it forward into the ML pipeline. 
As \jt, Weka has been mainly designed with in-memory\footnote{All the data available from input dataset is loaded in RAM memory before the processing task starts its own work.} batch-learning in mind, where all the input data is available before processing algorithms start using it. 
A minimal support to online learning with streaming data is also available, but the number of provided algorithms is very small and not comparable to variety and quantity of available batch learners.\footnote{See \url{http://weka.sourceforge.net/packageMetaData/} for a comprehensive list of ML algorithms available in Weka.}
Weka allows an easy access to its underlying functionalities by providing to its users several GUIs usable to quickly perform data mining tasks. 
These tools allow the user to visually define experiments by selecting data sources and filters to be applied, build supervised or unsupervised models, and graphically evaluate models and results obtained on the specific problem. 
On the other hand, \jt does not have an equivalent set of tools, but it relies only on the Java API in order to provide access to all its own features. 
Another important difference between our software and Weka is that \jt, while loading data from an input source during indexing phase, build an in-memory compressed index (see the IIndex structure in section \ref{jatecs:design_concepts}) over the input data allowing a) fast on-the-fly queries on the data; b) low memory consumption by exploiting the high sparseness of the textual data. 
In particular, the point a) allow to design ML algorithms conveniently and with few compromises in terms of performance, the point b) allow to work with datasets rather big even on a simple desktop machine. Lastly, another important aspect differentiating \jt from Weka is that out software provides ML algorithms which are text-domain specific and cover more advanced use cases than traditional data mining tasks (e.g. text quantification, cross\-lingual text classification, etc.).

Scikit-learn\cite{scikit-learn} is a popular open source Python module integrating a wide range of state-of-the-art ML algorithms for medium-scale supervised and unsupervised problems.\footnote{See \url{http://scikit-learn.org/stable/} for a complete list of what the software offers to developers.} 
The software is heavily used in the Python community because the API is simple and very well documented and the code is fast and optimized.\footnote{Scikit-learn is built over the Numpy and Scipy libraries. These libraries provide optimized data representation and fast operations on dense/sparse arrays and matrices, together with efficient algorithms for linear algebra.} 
The set of provided functionalities is very similar to Weka, covering all the typical phases of a data mining tool like dataset loading, data transformations, learning,  hyper-parameters optimization and models evaluation. 
Like Weka and differently from \jt, it is a generic data mining library so direct support for text analytics is very limited. 
For these purposes, Scikit-learn is often integrated with other Python libraries specialized in NLP tasks, e.g. NLTK library. 
Another difference from \jt is that Scikit-learn does not have an indexed representation of its input data for fast preprocessing data or build ML models with an higher level API. 
Instead it offers to developers a bunch of specialized transformers for data cleaning, dimensionality reduction, etc, that cover a lot of common cases. 
If required a developer can write custom transformers but the entire process is more complex if compared to the usage of an higher level API like the one provided by \jt. 
Finally, as Weka, Scikit-learn covers typical data mining scenarios but it is inadequate to deal with advanced NLP applications like textual transfer learning or text quantification.

Several libraries available in the market are specialized just on few specific aspects of a typical generic ML process. NLTK\cite{NLTK} and OpenNLP\cite{OpenNLPWebsite} are ML-based toolkits focused on NLP tasks. 
They provide solutions to most common text analytics problems like tokenization, sentence segmentation, part-of-speech tagging, named entity recognition, and so on, yet they lack a rich set of ML algorithms and tools like those provided by Weka and Scikit-learn. 
spaCy\cite{spaCy} is another notable Python NLP library focusing only on labelled dependency parsing, named entity recognition and part-of-speech tagging but providing extremely fast implementations (according to several benchmarks, the authors claim that their system is the fastest in the world) and top-class accuracy. 
\jt offers some of these NLP capabilities in its core library, anyway the framework design allows, in the case a specific feature is missing, to easily plug in a specific implementation of the wanted algorithm (or a wrapper around the specific tool) in its workflow. 

LibSVM\cite{libSVM} and SvmLight\cite{SvmLight} are very popular software packages providing in-memory fast implementations of various types of SVMs (Support Vector Machines), suitable for multiclass/binary classification and regressions and supporting several kernel types (linear, polinomial, RBF, etc.). 
Both these software have been integrated into \jt by wrapping their respective public interfaces. 
SHOGUN\cite{SHOGUN} is another machine learning toolbox focuses on large scale kernel methods, especially on Support Vector Machines. 
It provides similar algorithms as LibSVM and SvmLight but it offers an extended set of kernels including various recent string kernels like Fisher, TOP, Spectrum, etc. 
Moreover the software also implements a number of linear methods like Linear Discriminant Analysis (LDA), Linear Programming Machine (LPM), (Kernel) Perceptrons and features algorithms to train Hidden Markov models.

In the last years, with the data explosion due to increasing Internet popularity, several ML tools for BIG DATA processing have emerged in the market. 
MLlib\cite{MLlib} is a ML library built over Apache Spark which offers common ML and statistical algorithms. 
It includes components to perform basic statistics (correlations, stratified sampling, hypothesis testing, etc.), classification and regression (SVM, logistic regression, Naive Bayes, etc.), collaborative filtering, clustering (k-means, LDA, etc.), and other basic methods for feature extraction and transformation. 
As most of the other tools we have cited, it is a generic ML tool, so it does not provide algorithms targeting specifically NLP and text analytics. 
H$_{2}$O\cite{H2O} is another generic fast and scalable ML software suite which puts a lot of emphasis in the integration with other popular third party software (mainly Apache Spark and Apache Hadoop) and the possibility to use its features in several different ways (from Java, R, Python, Excel, etc.). 
It provides several common supervised and unsupervised learning methods (generalized linear modeling, distributed random forest, Naive Bayes, k-means, etc.) and a deep learning classifier/regressor based on an optimized distributed implementation of a classic feed forward neural network. 
Another generic ML software quite popular in the community is Apache Mahout\cite{Mahout}. 
It provides implementations of distributed or otherwise scalable machine learning algorithms focused primarily in the areas of collaborative filtering, clustering and classification. 
The software initially was implemented on top of Hadoop map-reduce, but in the last two years, thanks to the emerging of new fast data processing software like Apache Spark or Apache Flink, several algorithms have been ported to these new systems.         


\section{Conclusion}

\jt{} is a Java machine learning framework designed to support text mining processes.
It is highly modular, with the definition of interfaces for any information processing step of a typical ML pipeline, thus supporting a plug and play style exploration of different algorithms in a point of the pipeline with almost no adaptation required for the other components in the rest of the pipeline.

\jt{} also implements most of its interfaces through abstract classes, which provide a typical implementation of the basic functionalities of the interface, reducing the amount of code needed to be written when creating a new implementation of an interface. For example, the loops that apply a classification model to a set of documents, either document-by-document first or category-by-category first, are implemented in the Classifier class, and their rewriting for a new classification algorithm is needed only when none of these two methods of classification fits the way the classification algorithm works, a case we never faced.

A rich number of text processing methods are implemented, from tokenization and parsing, to weighting and feature selection.
Similarly, a rich number of ML algorithm are implemented or wrapped from state-of-the-art libraries.
\jt{} provides complete implementations to run the typical tasks of classification and clustering, as well as other tasks that are not support by most of the ML frameworks currently available, such as active learning, training data cleaning, transfer learning, and quantification.


\bibliographystyle{wileyj}
\bibliography{biblio}

\end{document}